\documentclass[journal]{IEEEtran}
\usepackage{graphicx}
\usepackage{amssymb,amsmath,bm}
\usepackage{textcomp}
\usepackage{algorithm}
\usepackage{algpseudocode}
\usepackage{amsmath}
\usepackage{graphics}
\usepackage{epsfig}
\usepackage{multirow}
\usepackage{amsmath}
\usepackage{url}
\usepackage{enumitem}
\usepackage{mathrsfs}
\usepackage{bm}
\usepackage{epstopdf}
\usepackage{array}
\usepackage{color}

\ifCLASSINFOpdf
\else
\fi
\begin{document}
%
\title{Sequence Discriminative Training for Deep Learning based Acoustic Keyword Spotting}
%
%
%
%

\author{Zhehuai~Chen,~\IEEEmembership{Student Member,~IEEE,}
        Yanmin~Qian,~\IEEEmembership{Member,~IEEE,} and
        Kai~Yu,~\IEEEmembership{Senior Member,~IEEE}

\thanks{This work was supported by the National Key Research and Development Program of China (Grant No.2017YFB1002102), the China NSFC project (No. 61603252), the China NSFC project (No. U1736202). Experiments have been carried out on the PI supercomputer at Shanghai Jiao Tong University.
}
\thanks{
The authors are with the Computer Science and Engineering Department, Shanghai Jiao Tong University, Shanghai 200240, China.
Corresponding authors are Kai Yu  and Yanmin Qian (email: \{kai.yu,yanminqian\}@sjtu.edu.cn).
}
}

\maketitle

\begin{abstract}
Speech recognition is a {\em sequence prediction} problem. Besides employing various deep learning approaches for frame-level classification, sequence-level discriminative training has been proved to be indispensable to achieve the state-of-the-art performance in large vocabulary continuous speech recognition (LVCSR). However, keyword spotting (KWS), as one of the most common speech recognition tasks, almost only benefits from frame-level deep learning due to the difficulty of getting competing sequence hypotheses. The few studies on sequence discriminative training for KWS are limited for fixed vocabulary or LVCSR based methods
and have not been compared to the state-of-the-art deep learning based KWS approaches. In this paper, a  sequence discriminative training framework is proposed for both fixed vocabulary and  unrestricted acoustic KWS. Sequence discriminative training for both sequence-level generative and discriminative models are systematically investigated. By introducing word-independent phone lattices or non-keyword {\em blank} symbols to construct competing hypotheses, feasible and efficient sequence discriminative training approaches are proposed for acoustic KWS. Experiments showed that the  proposed approaches obtained consistent and significant improvement in both fixed vocabulary and unrestricted KWS tasks, compared to previous frame-level deep learning based acoustic KWS methods. 




\end{abstract}



\IEEEdisplaynontitleabstractindextext

%
\IEEEpeerreviewmaketitle

\section{Introduction}\label{Sec:intro}


Keyword spotting (KWS) is one of the most widely used speech-related techniques, which requires a highly accurate and efficient recognizer specializing in the detection of some words or phrases of interest in continuous speech. KWS has broad applications, such as speech data mining~\cite{zhou2005data}, low resource audio indexing~\cite{shen2009comparison}, spoken document retrieval~\cite{garofolo2000trec} and wakeup-word recognition~\cite{chen2014small}. The last two applications are considered in this paper.

KWS techniques can be categorized into two groups: 
i) Unsupervised  {\em query-by-example} (QbyE) \cite{zhang2009unsupervised,barakat2012improved,chen2015query}, which utilizes keyword audio samples to generate  a set of keyword templates and matches them against testing audio samples to spot keywords.
ii) Supervised text-based method, which can be further divided into  {\em large vocabulary continuous speech recognition} (LVCSR)  based methods~\cite{garofolo2000trec,ng2000subword} and  {\em acoustic KWS} \cite{mandal2014recent}~\footnote{A  branch of newly proposed end-to-end methods~\cite{kintzley2011event,audhkhasi2017end} can also be viewed as a variant of it.}.
For LVCSR based methods, 
in training stage, a word or sub-word recognition system is constructed.
Acoustic and language models are used to transcribe speech into a database of text or lattice during testing stage.
Keyword searching is conducted on the database to get the final result.
Acoustic KWS
models the target keywords or sub-word sequences using an acoustic model without a language model. Some methods further include a series of non-keyword elements in the model~\cite{sukkar1996utterance}.
QbyE is mainly used in low
resource audio indexing, which is not the focus of this paper. 
In spoken document retrieval, LVCSR based methods often show better performance than acoustic keyword spotting based method. However, LVCSR based methods have some inevitable shortcomings: requirement of large vocabulary coverage in training dataset, large computational resource requirement  in both training and testing stage~\footnote{except for low resource speech recognition, e.g. Babel project~\cite{gales2014speech}. }, and out-of-vocabulary (OOV) problem, etc. These shortcomings limit its deployment in many practical applications such as wakeup-word recognition. Furthermore, LVCSR based KWS methods ignore the special characteristics  of KWS discussed in Section~\ref{Sec:kws-and-lvcsr}, and the performance improvements mainly rely on the advances of acoustic and language model in LVCSR. Therefore, this paper is focused on acoustic KWS.

In acoustic keyword spotting,  models are typically trained  to classify individual frames. Recent advances include two folds. First, applying a stronger frame-level classifier, deep neural network, yields significant improvements~\cite{chen2014small,sainath2015convolutional}.
Second, as speech recognition is inherently a sequence prediction problem, traditional GMM-HMM based systems achieve significantly better performance when trained using sequence discriminative criteria like discriminatively trained sub-word verification function~\cite{sukkar1996vocabulary}, minimum classification error (MCE)~\cite{sandness2000keyword} and performance-related discriminative training~\cite{keshet2009discriminative}. 
Recently, within the deep learning framework, word-based connectionist temporal classification (CTC) model has also been used for KWS~\cite{fernandez2007application}. 
In all above sequence discriminative training methods,  the complete search space modeling, i.e. hypothesis modeling,  is the key of the success.
However,  in KWS, the in-domain search space specified by keyword sequences is much smaller. Thus the out-of-domain search space should be modeled by specific non-keyword elements as competitors.
The difficulties in getting competing sequence hypotheses limit the usage of sequence discriminative training in KWS.
Especially in unrestricted KWS, the possible competing words are usually not enumerable and the competing hypotheses generation is computationally expensive if using the same procedure as in LVCSR~\cite{povey2005discriminative}. 

This paper proposes a sequence discriminative training framework for deep learning based unrestricted acoustic KWS. According to whether the model is defined for sequence conditional likelihood or sequence posterior probability, there are two types of sequence models: {\em generative sequence models} (GSM) such as HMM, and {\em discriminative sequence models} (DSM) such as CTC. For GSM, sequence discriminative training requires applying Bayes' theorem at sequence level to derive sequence conditional likelihood to posterior probability, while for DSM, sequence posterior probability can be used.  

For both frameworks, competing hypotheses handling is the key difficulty.
The paper proposes two methods to solve the problem: implicitly modeling  a
 sub-word level language model and explicitly modeling
 non-keyword symbols. 
In HMM, inspired by the success of applying a pruned phone level language model to replace the word lattices in LVCSR discriminative training~\cite{povey2016purely,chen2006advances},  the keyword sequences are modeled by a sub-word level acoustic model, and 
a corresponding language model is used to model the complete search space. 
To strengthen the discrimination ability of keywords,
their gradients  are weighted
more significantly than those on non-keywords. 
Moreover, various neural network architectures and discriminative training criteria are compared. 
In CTC, non-keyword model units are introduced explicitly. Namely, the search space of sub-word level CTC based KWS is composed of keywords,  phone boundaries ($\tt blank$) and word boundaries ($\tt wb$). Additional non-keyword spans ($\tt filler$) are introduced in word level CTC based KWS.
Lastly, an efficient post-processing algorithm is  proposed to include phone confusions in the hypothesis searching. 

The major contributions are summarized as follows:
i) The first
work to systematically investigate sequence discriminative training
for both generative and discriminative sequence models.
ii) Propose novel methods to construct competing hypotheses for sequence
discriminative training for acoustic KWS and significantly improve the
performance.
iii) Propose efficient post-processing methods to
include phone confusion in hypotheses search.

The rest of the paper is organized as follows. In Section~\ref{Sec:kws-and-disc},  the acoustic modeling in KWS is briefly reviewed.
In Section~\ref{Sec:sdt-review}, the traditional discriminative training methods are summarized.
In Section~\ref{Sec:kws-disc-proposed} and Section~\ref{Sec:kws-ctc}, the proposed sequence discriminative training methods for deep learning based KWS are introduced respectively in  CTC framework and HMM framework. Experiments are conducted on 
unrestricted KWS (spoken document retrieval task), and fixed vocabulary KWS (wakeup-word recognition task) in Section~\ref{Sec:exp}, followed by the conclusion in Section~\ref{Sec:conclu}.

\section{Acoustic Modeling for Keyword Spotting}
\label{Sec:kws-and-disc}

\subsection{Comparison between LVCSR and KWS}
\label{Sec:kws-and-lvcsr}

LVCSR and acoustic KWS are two related but different speech recognition tasks. LVCSR focuses on accurately transcribing of the whole utterance, whereas KWS focuses on detecting some specific words or phrases of interest. Although some common techniques can be shared by the two tasks, they have different requirements on acoustic modeling. To show that it is not trivial to apply the sequence discriminative training techniques (originally developed for LVCSR) to KWS, it is necessary to discuss the special requirements of acoustic modeling for KWS. 

\begin{itemize}
\item {\em Search space}. Due to extremely small vocabulary size, the in-domain search space of KWS is much smaller. 
Meanwhile, there are much more  non-keywords in KWS  than the out-of-vocabulary (OOV) words in LVCSR. Hence specific non-keyword models should be added into the search space of KWS system~\cite{sukkar1996utterance}\cite{sukkar1996vocabulary} to represent out-of-domain search space. 
\item {\em Model granularity}. Since the vocabulary in LVCSR is large,  acoustic model granularities smaller than word are usually used~\footnote{Recent progress in end-to-end system makes word or sub-word level modeling become competitive~\cite{graves2013speech,chan2016end,e2e-2018} and efficient~\cite{chen-is2016}. But the techniques have not been widely adopted.}, e.g., clustered tri-phones, which enhances both data efficiency and robustness~\cite{young1994state}.
However, there is no such consideration for KWS, thus the
model granularity can be keyword, sub-word, phone, tri-phone, etc.
\item {\em Decoding}. In LVCSR, decoding refers to the search process to 
find the most likely sequence of labels given acoustic and language models. In contrast, acoustic KWS usually does not require a language model but needs post-processing after the frame-level acoustic model inference. The post-processing method can be categorized into three groups: i) Posterior smoothing~\cite{chen2014small}. ii) Model based inference~\cite{cas-icassp17}. iii) $\tt filler$ based decoding~\footnote{In some recent works~\cite{chen2014novel,chen2017iscide}, a small language model can be applied in the $\tt filler$ modeling and shows moderate improvement.}. The first two groups aim to filter out the noise posterior output by heuristic or data-driven methods, respectively. The third group attempts to model the previously described out-of-domain search space, which will be explained in Section~\ref{Sec:post-process} in detail.
\end{itemize}


\begin{figure*}
  \centering
    \includegraphics[width=0.85\linewidth]{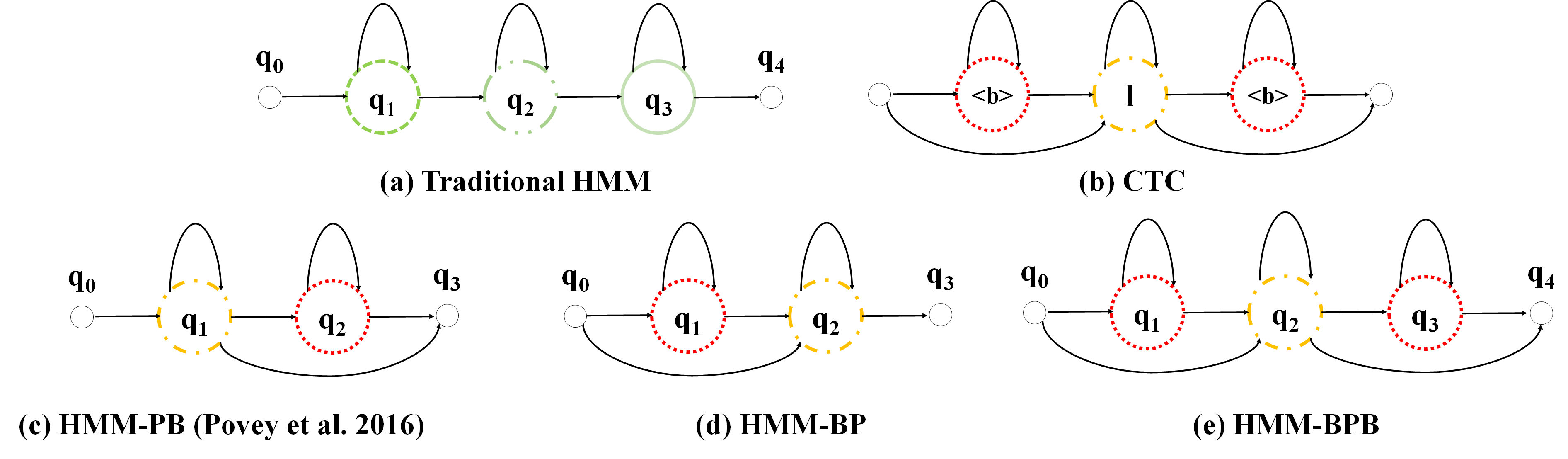}
    \caption{\it Illustration of the Hidden State Topologies in HMM, CTC and Proposed Methods. In the last three topologies, B refers to $\tt blank$ HMM state and P refers to the label output HMM state. Each  colored circle represents an HMM state with state occupation modeled by neural networks. 
    The dash-dot and orange states simulate the output label modeling, i.e. $\mathbf{l}$ in CTC. Each is assigned to a specific model unit. The dot and red states simulate the $\tt blank$ modeling, i.e. $\tt \langle b \rangle$ in CTC, and they are assigned to another model unit. The un-colored states don't consume feature vectors (non-emitting states). 
    The self-loop transition indicates that the transition model accepts repetition in the current state. We compare these topologies in Section~\ref{Sec:lfmmi-train}.}
    \label{fig:hmm-topo}
\end{figure*}

\subsection{Acoustic Modeling for KWS}
\label{Sec:am}
The acoustic keyword spotting based method are typically trained  to classify individual frames. 
In a deep learning based HMM hybrid system (NN-HMM) whose model granularity is the tri-phone state, a neural network is trained to calculate posterior probabilities of  HMM states.
Specifically, for an observation ${\bf o}_{ut}$ corresponding to time $t$ in utterance $u$,  $y_{ut}(s)=P(s|{\bf o}_{ut})$ is the output of the neural network for the HMM state $s$. The formulation is similar to traditional GMM-HMM based systems~\cite{young1994state}, except for the pseudo log-likelihood $\log p({\bf o}_{ut}|s)$ of HMM states $s$,
\begin{equation}
\label{equ:dnn-hmm-define}
\begin{split}
 p({\bf o}_{ut}|s) \propto \frac{ y_{ut}(s)}{ P(s)}
\end{split}
\end{equation}
where $P(s)$ is the prior probability of state $s$.  
In deep learning based system whose model granularity is a keyword, the posterior probability of the keyword $w$ is directly trained:
\begin{equation}
\label{equ:dnn-kw-define}
\begin{split}
P(w|{\bf o}_{ut})= y_{ut}(w)
\end{split}
\end{equation}
and the previously discussed post-processing method is applied on the frame level posterior probability $y_{ut}(w)$ in each utterance.

It is common to use the negative
log posterior as the cross-entropy objective function.
\begin{equation}
\label{equ:dnn-hmm-ce}
\begin{split}
\mathcal{F}_{\tt{CE}}=- \sum_{u} \sum_{t} \log y_{ut}(s^{(r)}_{ut})
\end{split}
\end{equation}
where $s^{(r)}_{ut}$ is the reference label at time $t$ for utterance $u$ obtained from a state level force alignment~\cite{woodland1994large}.

\section{Sequence Discriminative Training}
\label{Sec:sdt-review}
As speech recognition is inherently a sequence prediction problem, sequence level criteria are likely to improve the performance.
Depending on how the sequence model is defined, there are two types of sequence discriminative training approaches, one for generative sequence model (GSM) such as HMM,  and the other is for discriminative sequence model (DSM) such as CTC or encoder-decoder model. 


\subsection{Generative Sequence Model and Discriminative Sequence Model in ASR}
\label{Sec:sgm-and-sdm}

One interesting and important characteristics in speech recognition is the nature of variable-length acoustic features and label sequences. 
Sequence models are consequently needed to model the relationship between the two. 
Most sequence models usually consist of a component to model temporal characteristics, i.e. frame transition, and another frame-level 
component, e.g., GMM~\cite{woodland1994large} and neural networks~\cite{hinton2012deep}, to model the local characteristics of acoustic features~\footnote{The recent proposed discriminative sequence model, encoder-decoder~\cite{chan2016end}, directly operates at sequence level without sequence decomposition. But in KWS, both the performance and the runtime delay are still not satisfactory. Thus it is not included in the discussion below.}. 
All sequence discriminative training methods in this paper are based on  frame-wise decomposition.
Depending on whether conditional likelihood (or joint distribution) or posterior is modeled, a probabilistic model can be classified as {\em generative} or {\em discriminative}. It is worth noting that sequence level and frame level models can be either generative or discriminative. 
For example, the widely used hybrid NN-HMM model employs a generative sequence model with a discriminative frame-level classifier. 
In this paper, we are interested in sequence level models. A comparison of model structure and optimization framework between generative and discriminative sequence models is given below.


A {\em generative sequence model} is defined as the conditional likelihood $p(\mathbf{O}|\mathbf{L})$, where ${\mathbf O}$ is the feature observation sequence and $\mathbf L$ is the label sequence. Hidden Markov Model (HMM) is a typical example of generative sequence model~\footnote{Another GSM example is Kalman filtering based method~\cite{digalakis1991dynamical,abbeel2005discriminative}.}. 
In  NN-HMM hybrid systems, the  speech signal dynamics are modeled with HMMs and the
observation likelihoods are estimated using neural networks.
\begin{eqnarray}
\label{equ:hmm-model}
p(\mathbf{O}|\mathbf{L})&=&\!\!\!\!\!\!\sum_{\mathbf{q}\in\mathcal{A}(\mathbf{L})}\!\!\!p(\mathbf{O},\mathbf{q}|{\mathbf L}) =\sum_{\mathbf{q}}\prod_{t=1}^{T} p({\bf o}_{t}|q_t) P(q_t|q_{t-1})\nonumber\\
&\propto&\sum_{\mathbf{q}}\prod_{t=1}^{T} \frac{P(q_t|{\bf o}_{t})}{P(q_t)}P(q_t|q_{t-1})
\end{eqnarray}
where $\mathbf{L}$ is the label sequence, e.g., a word sequence in LVCSR. 
$\mathbf{q}$ is the HMM state sequence and  $q_t$ is the HMM state at frame $t$. $P(q_t|q_{t-1})$ is the HMM state transition probability and $P(q_t)$ is the state prior probability of $q_t$.
$\mathcal{A}$ is a mapping function from the label sequence $\mathbf{L}$, to its corresponding HMM state sequence $\mathbf{q}$,
\begin{equation}
\label{equ:a-func}
\begin{split}
\mathcal{A}:\mathbb{L}  \mapsto \{ q_0^{(1)},\cdots,q_4^{(1)},\cdots,q_4^{(|\mathbb{L}|)} \}
\end{split}
\end{equation}
$\mathbb{L}$ is the set of the label units in $\mathbf{L}$. $q_s^{(l)}$ is the s-th HMM state of the l-th HMM model.
Namely, each label unit, e.g. a tri-phone, corresponds to a  single HMM model.
Each of these HMM models contains five independent states, as shown in Figure~\ref{fig:hmm-topo}(a).
The state posterior $P(q_t|{\bf o}_{ut})$ is estimated by the neural network.
When a sequence level discriminative criterion, e.g. sequence posterior, is to be optimized, sequence level Bayesian decomposition has to be used to
allow the above generative model to be used with the discriminative criterion.

A {\em discriminative sequence model} is defined as $P(\mathbf{L}|\mathbf{O})$, the posterior probability of the sequence $\mathbf{L}$ given the feature sequence $\mathbf{O}$. Connectionist Temporal Classification (CTC)~\cite{graves2006connectionist} is one implementation of discriminative sequence model. It introduces a $\tt blank$ symbol to model the cross-label confusion~\footnote{Another DSM example is RNN transducer~\cite{graves2012sequence}, which allows the model to predict {\em null} outputs.}. Namely, the model always infers $ \tt blank$ symbol between $l_{i-1}$ and $l_{i}$.

\begin{equation}
\label{equ:ctc-model}
\begin{split}
P(\mathbf{L}|\mathbf{O})=\sum_{\mathbf{q}\in\mathcal{B}(\mathbf{L})}P(\mathbf{q}|\mathbf{O}) =\sum_{\mathbf{q}}\prod_{t=1}^{T} p(q_t|\mathbf{O})
\end{split}
\end{equation}
where $\mathcal{B}$ is a one-to-many mapping~\footnote{The original formulation in~\cite{graves2006connectionist} uses a many-to-one function and hence an inverse mapping function notation $\mathcal{B}^{-1}(\cdot)$ is used. Note that here we change the notation and use a one-to-many function to be consistent with Equation~(\ref{equ:a-func}) in CTC.}:
\begin{equation}
\label{equ:ctc-b}
\begin{split}
\mathcal{B}:\mathbb{L}   \mapsto  \mathbb{L} \cup \{\tt blank\}
\end{split}
\end{equation}
$\mathcal{B}$ determines the label sequence $\mathbf{L}$ and its corresponding set of model unit sequences $\mathbf{q}$. The mapping is done by inserting an optional and self-loop $ \tt blank$ symbol between each label unit $l$ in $\mathbf{L}$, as shown in Figure~\ref{fig:hmm-topo}(b). We compare different topologies in Section~\ref{Sec:lfmmi-train}. $P(q_t|\mathbf{O})$ is estimated by the neural network taking the feature sequence $\mathbf{O}$ as the input, e.g. long short term memory (LSTM)~\cite{hochreiter1997long}.
When a sequence discriminative model is used, the optimization of sequence discriminative criterion is straightforward. 

%

\subsection{HMM based Sequence Discriminative Training}
\label{Sec:sgm-sdt-intro}





Given hidden markov model (HMM), to construct a sequence discriminative training criterion, it is necessary to calculate the sequence posterior probability using Bayes' theorem,
\begin{equation}
\label{equ:map-dec}
\begin{split}
P(\mathbf{W}_u|\mathbf{O}_u)=\frac {p(\mathbf{O}_u|\mathbf{W}_u)P(\mathbf{W}_u)}{p(\mathbf{O}_u)}  
\end{split}
\end{equation}
Here, $\mathbf{W}_u$ is the word sequence of utterance $u$. $P(\mathbf{W}_u)$ is the language model probability. In KWS, $P(\mathbf{W}_u)$ is defined by  the prior probability of  keyword sequences and non-keyword elements.
$p(\mathbf{O}|\mathbf{W})$ is the corresponding acoustic part and can be obtained by,  
\begin{equation}
\label{equ:lexicon}
\begin{split}
p(\mathbf{O}|\mathbf{W})=\sum_{\mathbf{L}\in\mathcal{L}(\mathbf{W})} p(\mathbf{O}|\mathbf{L})P(\mathbf{L}|\mathbf{W})
\end{split}
\end{equation}
where $p(\mathbf{O}|\mathbf{L})$ is given by (\ref{equ:hmm-model}) from HMM. $\mathcal{L}$ is the mapping function from the word sequence $\mathbf{W}$ to its label sequence $\mathbf{L}$ of the sequence model discussed in Section~\ref{Sec:sgm-and-sdm}, e.g. tri-phone sequence in LVCSR. $P(\mathbf{L}|\mathbf{W})$ is the pronunciation probability~\cite{chen2015pronunciation} and usually decided by lexicons and language models.

The marginal probability $p(\mathbf{O})$ of the feature sequence $\mathbf{O}_u$, is modeled by the summation of the probability over all possible hypothesis sequences. 
\begin{equation}
\label{equ:po-prob}
\begin{split}
p(\mathbf{O}_u)=\sum_\mathbf{W} p(\mathbf{O}_u,\mathbf{W})= \sum_\mathbf{W} P(\mathbf{W}) p(\mathbf{O}_u|\mathbf{W})
\end{split}
\end{equation}
Here, $\mathbf{W}$ denotes one of the competing hypotheses, which are usually represented as a path in the decoding lattice.
As an example of the sequence discriminative training criteria, the maximum mutual information (MMI)~\cite{bahl1986maximum} is defined as below.
\begin{equation}
\label{equ:lvcsr-mmi}
\begin{split}
\mathcal{F}_{\tt{MMI}}
=\sum_{u} \log \frac {p(\mathbf{O}_u|\mathbf{W}_u)^{\kappa}P(\mathbf{W}_u)}{\sum_{\mathbf{W}} p(\mathbf{O}_u|\mathbf{W})^{\kappa}P(\mathbf{W})}  
\end{split}
\end{equation}
where the distribution, $P(\mathbf{W}_u|\mathbf{O}_u)$, in (\ref{equ:map-dec}), is scaled with the factor $\kappa$. 
Given the sequence posterior probability representation as in Equation (\ref{equ:lvcsr-mmi}), more advanced sequence discriminative training criteria can be derived within the Bayesian risk framework, which will be discussed in Section~\ref{Sec:lfmmi-train}.

Deep learning based LVCSR and LVCSR-based KWS can both be improved by sequence discriminate training~\cite{vesely2013sequence,meng2016non}. 
The sequence level competing hypothesis $\mathbf{W}$ in (\ref{equ:po-prob})
can be obtained by searching with a language model~\footnote{except for some grammar or keyword based recognition, where the language model is unavailable~\cite{sukkar1996utterance}.}.
Decoding lattice, as a compact approximation of the complete search space, is used to constrain the number of $\mathbf{W}$ to calculate  $p(\mathbf{O}_u)$  in Equation (\ref{equ:po-prob}) .

In acoustic KWS based methods, however, only the keyword sequence is defined, while the non-keyword competing hypothesis is unknown. The likelihood ratio based hypothesis testing framework~\cite{sukkar1996utterance} is proposed to  conduct discriminative training by penalizing the likelihood of the composite alternate hypotheses. The likelihood is modeled by two specific model units: $\tt filler$ model, $p(\mathbf{O}_u|\Phi)$ for non-keyword speech, and anti-keyword model, $p(\mathbf{O}_u|\Psi)$ for mis-recognitions. Compared with equation~(\ref{equ:lvcsr-mmi}),  $p(\mathbf{O}_u|\mathbf{W}_u)$ for keyword sequences is ignored and the logarithmic marginal probability is defined as below,
\begin{equation}
\label{equ:wbmve-po}
\begin{split}
\log p(\mathbf{O}_u)=\left\{\frac{1}{2}[\log p(\mathbf{O}_u|\Psi)^\lambda + \log p(\mathbf{O}_u|\Phi)^\lambda]\right\}^{1/\lambda}
\end{split}
\end{equation}
where the sequence level competing hypothesis $\mathbf{W}$ is generated from N-best recognition results.
In these methods, artificially dividing  non-keyword elements into two groups is imperfect. 
The pronunciation and acoustic environment differences exist within the non-keyword units. 
Moreover, these methods also fail to model the context  between keyword and non-keyword sequences. 
Furthermore, generating hypotheses from N-best recognition results is insufficient.
Therefore,  the modeling effect is deteriorated.
Finally, the method is not applicable for unrestricted KWS because the non-keyword elements cannot be defined in the training stage.

\subsection{CTC based Sequence Discriminative Training}
\label{Sec:sdm-sdt-intro}

{\em{Discriminative sequence  model}}  directly calculates the posterior
probability of the complete label sequence given the feature sequence as in (\ref{equ:ctc-model}).
Connectionist temporal classification (CTC) is an example, which operates at sequence level  to label unsegmented data using a sequence discriminative criterion~\cite{graves2006connectionist,huang2018ctc}. 
In~\cite{fernandez2007application}, CTC is applied to KWS.
The outputs of CTC are interpreted as a probability distribution over all possible keyword sequences $\mathbf{W}$. The objective function is defined as:
\begin{equation}
\label{equ:ctc-formu}
\begin{split}
\mathcal{F}_{\tt{CTC}}=\sum_{u} \log P(\mathbf{W}_u|\mathbf{O}_u)\\
=\sum_{u} \log \sum_{\mathbf{L}\in\mathcal{L}(\mathbf{W}_u)} P(\mathbf{L}|\mathbf{O}_u)P(\mathbf{W}_u|\mathbf{L})
\end{split}
\end{equation}
where $\mathbf{L}$ is the label sequence of the CTC model, e.g. sub-word sequence. Practically, mono-phone is taken as the sub-word unit.
In acoustic KWS, there is no language model. Thus $P(\mathbf{W}_u|\mathbf{L})$ is effectively a deterministic mapping given the lexicon and the keyword sequences.
$P(\mathbf{L}|\mathbf{O}_u)$ can be further mapped to CTC label sequence and factorized into frame-level, as shown in Equation (\ref{equ:ctc-model}).
Note that an extra $\tt blank$ unit is introduced at frame level in equation  (\ref{equ:ctc-model}) to model confusion spans of the speech signal.
The sequence discriminative criterion is the summation of the posterior probabilities of all  possible CTC label sequences.

In word level CTC~\cite{fernandez2007application}, 
although it is naturally a sequence level criterion, it does not model the non-keyword elements directly. Namely, $\tt blank$
is inserted between keywords to model the context between them.
Therefore, the sequence level criterion improves the sequence prediction ability between keywords but not between keywords and non-keywords.

\section{Keyword Spotting Using HMM Based Sequence Discriminative Training}
\label{Sec:kws-disc-proposed}

As discussed in Section~\ref{Sec:sgm-sdt-intro}, the main difficulty for HMM based sequence discriminative training is to generate word-level competing hypotheses. Inspired by the  recent success in applying a  pruned phone level language model to replace  word lattices in discriminative training in LVCSR (referred to as {\em lattice-free maximum mutual information} (LF-MMI)~\cite{povey2016purely,chen2006advances}), a general sequence discriminative training framework is proposed for unrestricted KWS. Here, the keyword sequence is modeled by sub-word level acoustic models. Correspondingly,
a language model over all sub-word units is used to model the complete search space, namely 
keyword sequences and sequence level competing hypotheses. 
In this framework, since only sub-word unit language model is used to generate competing hypotheses, dealing with non-keywords within sequence discriminative training framework becomes feasible and effective. 

\subsection{Model Training}
\label{Sec:lfmmi-train}

In the proposed sub-word level acoustic model, (\ref{equ:lvcsr-mmi}) is transformed to (\ref{equ:kws-mmi}) denoted as $\tt{LF\text{-}MMI}$.
\begin{equation}
\label{equ:kws-mmi}
\begin{split}
\mathcal{F}_{\tt{LF\text{-}MMI}}
=\sum_{u} \log \frac {\sum_{\mathbf{L}_u} p(\mathbf{O}_u|\mathbf{L}_u)^{\kappa}P(\mathbf{L}_u)}{\sum_{\mathbf{L}} p(\mathbf{O}_u|\mathbf{L})^{\kappa}P(\mathbf{L})}  
\end{split}
\end{equation}
where $\mathbf{L}$ is the sub-word level sequence, e.g. phone sequence. $\mathbf{L}_u$ is the reference label sequence and it is sub-word level.  
$p(\mathbf{O}_u|\mathbf{L})$ and $p(\mathbf{O}_u|\mathbf{L}_u)$ are obtained by Equation~(\ref{equ:hmm-model}).

Compared with Equation (\ref{equ:lvcsr-mmi}), there are several differences in this framework~\cite{povey2016purely}: 
i) The reference label sequences $\mathbf{L}_u$ used in the numerator of Equation (\ref{equ:kws-mmi}) have multiple candidates because it is the soft transcript alignment with the left and right frame shift window.  Hence the summation of all possible alignments is taken in the nominator. 
ii) The competing hypothesis sequences in denominator and the probability $P(\mathbf{L})$, $P(\mathbf{L}_u)$  are estimated by a sub-word level language model trained on the training transcription.  
iii) A specific HMM topology is proposed to model each  tri-phone by two states, referred to as HMM-PB in Figure~\ref{fig:hmm-topo}(c). Specifically, the state ${\bf q}_2$ simulates the $\tt blank$ model in CTC,  $\tt \langle b \rangle$ in Figure~\ref{fig:hmm-topo}(b), while the other state ${\bf q}_1$ simulates the output label unit,  $\mathbf{l}$ in Figure~\ref{fig:hmm-topo}(b). The difference is that each tri-phone in \cite{povey2016purely} keeps its own version of $\tt blank$.
iv) the output frame rate is subsampled by 3 folds.

To better apply $\tt{LF\text{-}MMI}$ in KWS, several improvements are made in this paper: 
i) The model unit is mono-phone rather than tri-phone.
Firstly, in model inference, efficiency is greatly improved due to fewer model units.  Secondly, in the post-processing discussed in Section~\ref{Sec:post-process}, the $\tt filler$ construction also becomes much simpler, which improves both efficiency and robustness.
ii) The HMM topology~\footnote{In the original HMM topology in HTK~\cite{woodland1994large}, the first states are non-emitting states, and they are not assigned to any model unit, which can directly transit to next states.  In Kaldi~\cite{povey2011kaldi}, The first states are emitting states. To make it consistent, we always use non-emitting first states in explaining.} is changed to Figure~\ref{fig:hmm-topo}(d-e), inspired by the CTC topology,  Figure~\ref{fig:hmm-topo}(b). Specifically, 
CTC allows $\tt blank$ to exist before and after the label output. Repeated symbols are also allowed. 
In HMM-PB, $\tt blank$ (state $\mathbf{q}_2$ in Figure~\ref{fig:hmm-topo}(c)) exists only after the label output (state $\mathbf{q}_1$).
In light of the performance improvement gained from label delay~\cite{amodei2015deep}, the proposed HMM-BP structure delays the phone label outputs and infers $\tt blank$ before the label output. HMM-BPB is proposed  as a complete simulation to CTC except the phone-independent $\tt blank$ states~\footnote{
Sharing the $\tt blank$ state across phone in Figure~\ref{fig:hmm-topo}(e) shows worse performance than Figure~\ref{fig:hmm-topo}(e). We believe, at least in small dataset, the $\tt blank$ modeling is always the bottleneck. A global $\tt blank$ needs sufficient data to model all the confusion spans between  different kinds of phones. 
}.
The states simulating $\tt blank$ modeling in each mono-phone HMM are further bound together as depicted in Figure~\ref{fig:hmm-topo}(e). Thus, all the above topologies, HMM-PB, HMM-BP and HMM-BPB, require two model units respectively.
Experimental comparison will be given in Section~\ref{Sec:exp-model-arch}.

To further improve the discrimination ability, a number of advanced sequence discriminative training criteria are investigated for KWS in this paper. First, the likelihoods of competing hypothesis sequences that contain more errors are boosted~\cite{povey2008boosted}, called boosted  $\tt{ LF\text{-}MMI}$, $\tt{ LF\text{-}bMMI}$.
\begin{equation}
\label{equ:kws-bmmi}
\begin{split}
\mathcal{F}_{\tt{LF\text{-}bMMI}}
=\sum_{u} \log \frac {\sum_{\mathbf{L}_u} p(\mathbf{O}_u|\mathbf{L}_u)^{\kappa}P(\mathbf{L}_u)}{\sum_{\mathbf{L}} p(\mathbf{O}_u|\mathbf{L})^{\kappa}P(\mathbf{L})e^{-b\ \mathop{\max}_{\mathbf{L}_u} A(\mathbf{L},\mathbf{L}_u)}}  
\end{split}
\end{equation}
where $A(\mathbf{L},\mathbf{L}_u)$ is the state level accuracy of sequence $\mathbf{L}$ versus the reference label sequence $\mathbf{L}_u$,
$b$ is the boosting factor.
Because $\mathbf{L}_u$ used in the numerator is the soft alignment of the transcript, the boosted value is obtained by the best accuracy.
Another branch of discriminative training methods aims to minimize the expected error corresponding to different granularity of labels~\cite{gibson2006hypothesis}. The lattice-free state level minimum
Bayes risk ($\tt{LF\text{-}sMBR}$) is investigated in this work.

\begin{equation}
\label{equ:kws-smbr}
\begin{split}
\mathcal{F}_{\tt{LF\text{-}sMBR}}
=\sum_{u}  \frac {\sum_{\mathbf{L}} p(\mathbf{O}_u|\mathbf{L})^{\kappa}P(\mathbf{L})\mathop{\max}_{\mathbf{L}_u} A(\mathbf{L},\mathbf{L}_u)}{\sum_{\mathbf{L}} p(\mathbf{O}_u|\mathbf{L})^{\kappa}P(\mathbf{L})}  
\end{split}
\end{equation}


The proposed sequence discriminative training method can also be extended to fixed vocabulary KWS tasks, whose acoustic model is dependent on pre-defined keywords.
To strengthen the discrimination ability of the acoustic model for specific keywords,
the keyword related gradients are weighted
more significantly than those of non-keywords. 
A per-frame non-uniform weight can be added into the loss functions in Equation (\ref{equ:kws-mmi}-\ref{equ:kws-smbr}) similar to \cite{meng2016non}, which operates in MCE. The key point is to emphasize  the loss during the span of possible keyword false rejection and false alarm in the training data.
For example, one can make a non-uniform version of LF-MMI, referred to in the following as non-uniform LF-MMI ($\tt{NU\text{-}LF\text{-}MMI}$):
\begin{equation}
\label{equ:kws-nummi}
\begin{split}
\frac{\partial \mathcal{F}_{\tt{NU\text{-}LF\text{-}MMI}}}{\partial\log p({\bf o}_{ut}|s)}
=\frac{\partial \mathcal{F}_{\tt{LF\text{-}MMI}}}{\partial\log p({\bf o}_{ut}|s)}\cdot  \ell(t,u)
\end{split}
\end{equation}
where $s$ is the model unit and $p({\bf o}_{ut}|s)$ is the neural network output of state $s$ at frame $t$ in utterance $u$.
$\ell(t,u)$ is the derivative weight function given at frame $t$ in utterance $u$. $\ell(t,u)$ is defined as below.
\begin{equation}
\label{equ:deriv-weight}
\ell(t,u)=
\begin{cases}
\mathop{\min(\alpha,\beta)}& r_{ut}\in \mathbf{K}\ \land\ i_{ut}\in \mathbf{K}\\
\alpha& r_{ut}\in \mathbf{K}\ \land\ i_{ut}\notin \mathbf{K}\\
\beta& i_{ut}\in \mathbf{K}\ \land\ r_{ut}\notin \mathbf{K}\\
1& others
\end{cases}
\end{equation}
where $\mathbf{K}$ is the set of sub-word level keyword sequences. $r_{ut}$   is the sub-word level reference label at frame $t$ in utterance $u$, and $i_{ut}$ is the corresponding inference. $\alpha$ and $\beta$ are the boosting factors (larger than $1$) for the keyword false rejection and false alarm, respectively. For the first case in Equation (\ref{equ:deriv-weight}), $\mathop{\min(\alpha,\beta)}$ is taken because the model may have already inferred the keywords. $\ell(t,u)$ can be decided by traversing the training set with an initial acoustic model trained by LF-MMI.

\subsection{Post Processing}
\label{Sec:post-process}

As discussed in Section~\ref{Sec:kws-and-lvcsr}, post-processing methods can be divided into three groups. In this work, the posterior smoothing and $\tt filler$ based decoding are used.

\subsubsection{Posterior Smoothing}
\label{Sec:post-smooth}

The posterior smoothing method aims to filter out the noisy posterior output by a simple, yet effective approach. The method is originally proposed in~\cite{chen2014small} and can be summarized as follows. 
\begin{equation}
\label{equ:post-smooth-conf}
\begin{split}
P'(s'|{\bf o}_{ut'})=\mathcal{N}_1\left (\mathbf{P}(s|{\bf o}_{ut})^{(s=s',\ t\in [t'-\frac{1}{2}\mathrm w_{s},t'+\frac{1}{2}\mathrm w_{s}])}\right ) 
\end{split}
\end{equation}
\begin{equation}
\label{equ:post-smooth-conf2}
\begin{split}
P''(s'|{\bf o}_{ut'})=\mathcal{N}_2\left (\mathbf{P}'(s|{\bf o}_{ut})^{(s=s',\ t\in [t'-\mathrm w_{m}+1,t'])}\right ) 
\end{split}
\end{equation}
\begin{equation}
\label{equ:post-smooth-conf3}
\begin{split}
\mathcal C(\mathbf{k})^{(t')}=\mathcal{N}_3\left (\mathbf{P}''(s|{\bf o}_{ut})^{(s\in \mathbf{k},t=t')}\right ) 
\end{split}
\end{equation}
where $P(s'|{\bf o}_{ut'})$, $P'(s'|{\bf o}_{ut'})$ and $P''(s'|{\bf o}_{ut'})$ are three versions of the normalized posterior of the neural network model unit $s'$ at frame $t'$ in utterance $u$.  $\mathbf{P}(s|{\bf o}_{ut})^{(s\in \mathbf{s},t\in\mathbf{t})}$, $\mathbf{P}'(s|{\bf o}_{ut})^{(s\in \mathbf{s},t\in\mathbf{t})}$ and $\mathbf{P}''(s|{\bf o}_{ut})^{(s\in \mathbf{s},t\in\mathbf{t})}$ are three versions of the posterior sequence whose state $s$ belongs to set $\mathbf{s}$ and whose frame $t$ belongs to set $\mathbf{t}$. $\mathcal{N}_1(\cdot)$, $\mathcal{N}_2(\cdot)$ and $\mathcal{N}_3(\cdot)$ are three smoothing functions. In~\cite{chen2014small},  arithmetic mean, maximum function and geometric mean are used respectively.

Posteriors from the neural network are smoothed twice by fixed time windows of size $\mathrm w_{s}$ and $\mathrm w_{m}$ iteratively in (\ref{equ:post-smooth-conf}) and (\ref{equ:post-smooth-conf2}). The confidence $\mathcal C(\mathbf{k})^{(t')}$ of certain keyword sequence $\mathbf{k}$ at frame $t'$  is obtained by further smoothing the $P''(s'|{\bf o}_{ut'})$ inner the sub-word sequence of $\mathbf{k}$. $\mathbf{k}$ at frame $t'$ is then compared with keyword-dependent threshold to decide whether to spot it. 

In fix vocabulary KWS system, the threshold is keyword-dependent, which is discussed in Section~\ref{Sec:exp-sp-docu-detri-setup}.
In unrestricted KWS, whose acoustic model is independent of the pre-defined keywords, threshold estimation is defined as.
\begin{equation}
\label{equ:post-smooth-thres}
\begin{split}
P'(s'|{\bf o})=\mathcal{N}_1\left (\mathbf{P}(s|{\bf o})^{(s=s')}\right )
\end{split}
\end{equation}
\begin{equation}
\label{equ:post-smooth-thres2}
\begin{split}
\mathcal T(\mathbf{k})=\mathcal{N}_3\left (\mathbf{P}'(s|{\bf o})^{(s\in \mathbf{k})}\right ) 
\end{split}
\end{equation}
where $\mathbf{P}(s|{\bf o})^{(s\in \mathbf{s})}$ and $\mathbf{P}'(s|{\bf o})^{(s\in \mathbf{s})}$ are the posteriors whose state $s$ belongs to set $\mathbf{s}$. $\mathbf{P}(s|{\bf o})^{(s=s')}$ of each state $s'$ is obtained from state level forced-alignment over a development set. $P'(s'|{\bf o})$ is the normalized posterior estimation over all forced-aligned states $s'$. The threshold of a certain unrestricted keyword sequence $\mathbf{k}$ is obtained from its sub-word sequence as in (\ref{equ:post-smooth-thres2}). 

\subsubsection{$\tt filler$-based Decoding}
\label{Sec:fil-dec}

\begin{figure}
  \centering
    \includegraphics[width=\linewidth]{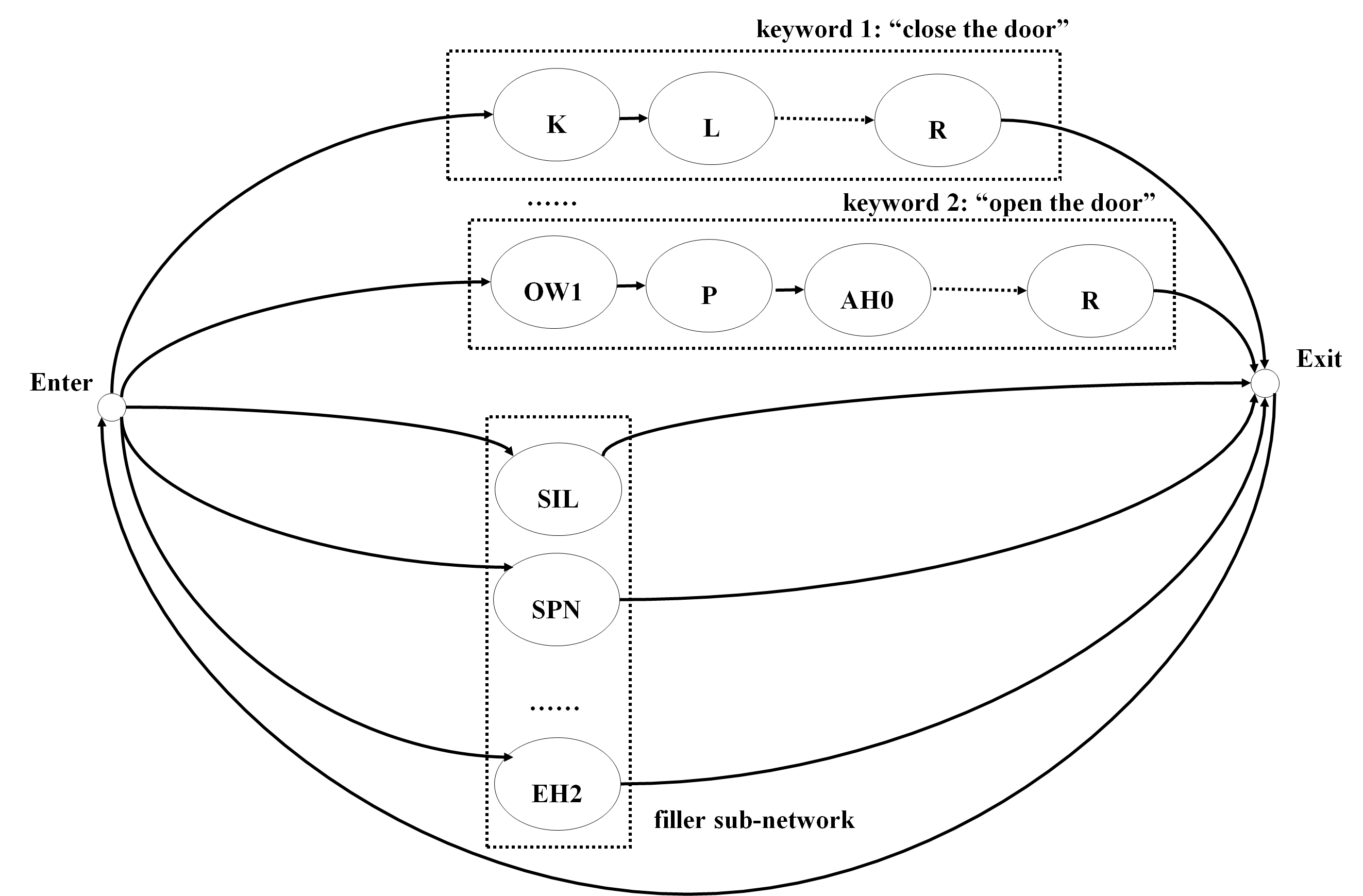}
    \caption{\it Illustration of the Search Space in the $\tt filler$ Based Decoding.}
    \label{fig:filler-graph}
\end{figure}

$\tt filler$ based decoding attempts to model the previously described out-of-domain search space by $\tt filler$. The search space in the proposed method is depicted in Figure~\ref{fig:filler-graph}.

It consists of two parts: the in-domain search space and the out-of-domain search space. The former is modeled by the sub-word sequences of the keywords, e.g., phone sequences in Figure~\ref{fig:filler-graph}. A $\tt filler$ sub-network is used to model the out-of-domain search space. $\tt filler$ sub-network is composed of all sub-word loops, each with a $\tt filler$ weight to trade off between the keyword false rejection and false alarm. 

\section{Keyword Spotting Using CTC Based Sequence Discriminative Training}
\label{Sec:kws-ctc}



\subsection{Model Training}
\label{Sec:modeltrain}

As discussed in \ref{Sec:sdm-sdt-intro}, the key of the success in sequence discriminative training is the competing hypothesis modeling.
In this paper, two directions are considered: word level modeling and sub-word level modeling respectively.

To better model competing hypotheses in word level CTC models,
a new model unit is introduced for non-keywords and the contexts between words. 
Similar to traditional acoustic KWS based methods, a non-keyword unit, $\tt filler$, is introduced and added into the CTC output label set. During the training stage, $\tt filler$ replaces non-keywords in the transcription. 

Another direction is to employ sub-word level models. 
The motivation includes two folds:
i) Any sub-word sequences, including non-keyword sequences, can be simultaneously modeled using sub-word models.
ii) The acoustic model can be independent of keywords, which is particularly useful for the unrestricted KWS task.
Moreover, 
a special label $\tt wb$ is introduced to model
the word boundaries in word level CTC model as in~\cite{zhuang-is2016}. In sub-word level CTC (phone CTC is taken in this work), $\tt wb$ and $\tt blank$ are used to model the word and phone boundaries  respectively. Distinguishing the type of the span in confusion modeling, can both improve the modeling effect and the post-processing discussed in the next section.
Furthermore, $\tt wb$ can deal with the case that the sub-word sequence of a certain short keyword is a substring of a longer keyword or non-keyword.

The formulation of introducing an additional competing hypothesis model is as follows,
\begin{equation}
\label{equ:ctc-kw}
\begin{split}
P(\mathbf{L}_u|\mathbf{O}_u)=P(\mathbf{L}_u'|\mathbf{O}_u)_{\mathbf{L}_u' = \mathcal{D}(\mathbf{L}_u)}
\end{split}
\end{equation}
$\mathcal{D}$ is a label mapping function. There are two forms of $\mathcal{D}$ defined for word level and sub-word level CTC respectively.
\begin{equation}
\label{equ:ctc-d}
\begin{split}
\mathcal{D}_{\tt{word}}:\mathbb{L}  \mapsto \mathbb{L}  \cup \{{\tt filler}\}\\
\mathcal{D}_{\tt{sub-word}}:\mathbb{L}  \mapsto \mathbb{L}  \cup \{{\tt wb}\}
\end{split}
\end{equation}
With the proposed label mapping and assuming output label independence, (\ref{equ:ctc-formu}) gives the CTC formulation, which can be optimized efficiently with the forward-backward algorithm~\cite{graves2006connectionist}.
Thus both in word level and sub-word level modeling,
the search space is composed of keywords, non-keywords, phone boundaries and word boundaries.

In ASR, HMM based sequence discriminative training is applied on CTC variants and achieves further improvement~\cite{sak2015fast,nict-icassp17}. We include these trials in unrestricted KWS in Section~\ref{Sec:exp-perf-comp}.

\subsection{Post Processing}
\label{Sec:post-process-ctc}

Besides the methods discussed in Section~\ref{Sec:post-process}, 
an efficient and concise {\em minimum edit distance} (MED) variant
post-processing algorithm for CTC trained systems is applied in test stage. 
The  inference posteriors from CTC model are always peaky. To get better sequence level competing hypotheses, a confusion matrix of phone deletion, insertion and substitution prior probability is estimated and integrated into the hypothesis searching.  
Inspired by \cite{chaudhari2007improvements}, the integration of phone confusion is based on CTC lattice and MED algorithm~\cite{7736093}. 

\begin{figure}[htbp!]
\centering
\includegraphics[width=\linewidth]{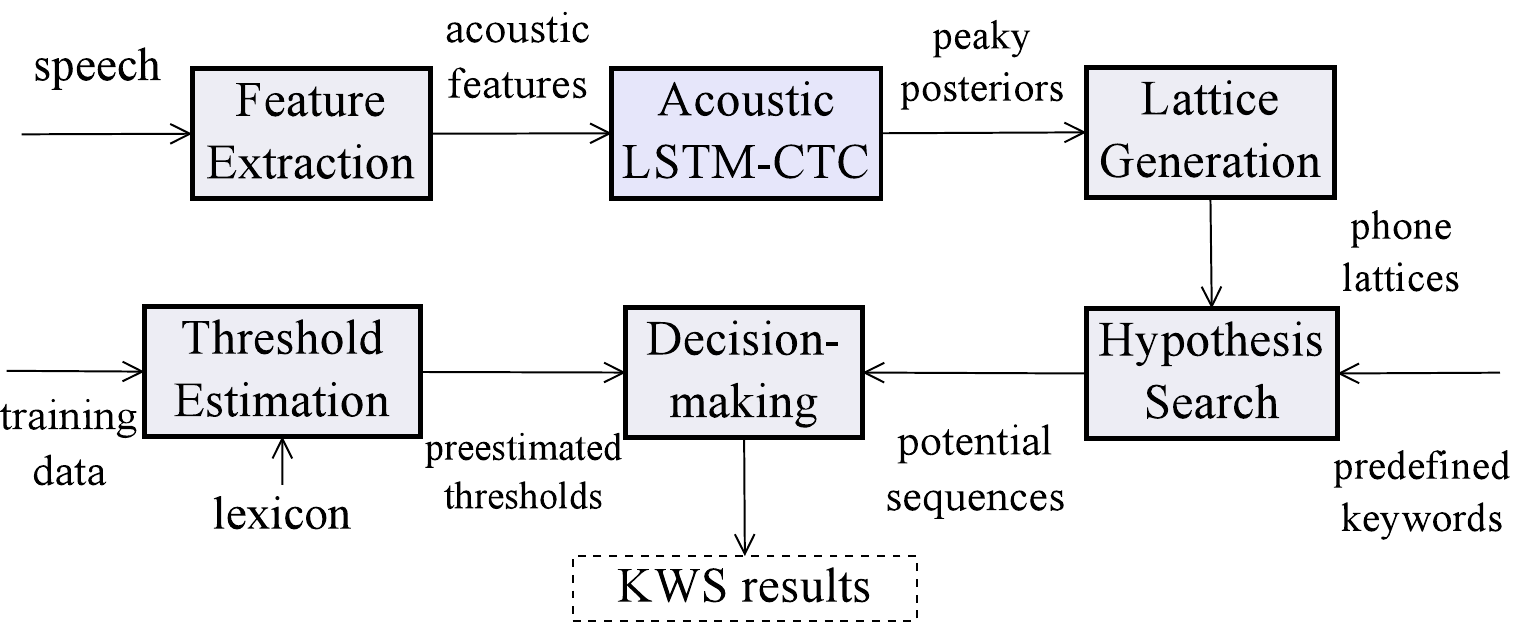}
\caption{\it Framework of the Proposed MED Method. The sub-word level CTC is taken as an example, namely the phone level CTC.}
\label{fig:med-framework}
\end{figure}

Figure~\ref{fig:med-framework} shows the framework of the MED method in the context of the sub-word CTC. 
In a speech utterance, 
the probability of certain keyword existing in the CTC lattice 
is estimated through multiplying the probabilities of each edit operation: insertion, deletion and substitution.
The probability of each operation is obtained by the MED between each hypothesis sequence and the keyword sequence,
Practically, for different phones, CTC model may have different modeling effects. Therefore, the threshold of a keyword should be dependent on its phone sequence. Prior statistics can be used to estimate an optimal contribution factor for each phone on a development set~\cite{zhuang-is2016}.

We also experiment with the introduction of phone confusion based MED method into HMM. 
However,  some problems  exist and the preliminary trial is not successful: i) In CTC, the MED is conducted on the proposed CTC lattice. The inference distribution of HMM is not as peaky as that of CTC~\cite{rosenberg2017end}. Thus the computational complexity is much larger. ii) In HMM, the neural network output $p({\bf o}_{ut}|q_t)$ is already combined with other knowledge sources in sequence discriminative training as  Equation (\ref{equ:hmm-model}) and (\ref{equ:map-dec}). Thus the improvement from phone confusion is insignificant.

\subsection{Comparison of Sequence Discriminative Training between HMM Framework and CTC Framework}
\label{Sec:disc-and-ctc}

In this paper, sequence discriminative training is classified into two categories according to the type of the sequence model: generative sequence model versus discriminative sequence model. It is then of interest to compare the CTC formulation (\ref{equ:ctc-formu}) in this section with the HMM formulation (\ref{equ:kws-mmi})  in Section~\ref{Sec:kws-disc-proposed}. The differences are summarized: 

\begin{itemize}
  \item {\em Generative and discriminative models}. In HMM, Bayesian approaches are
  used to provide temporal state transition probability, and  the observation likelihood is estimated through the neural network. In CTC, the posterior probability of  the label sequence given the feature sequence is directly modeled by the neural network.
  \item {\em Sequence modeling}. By introducing $\tt{blank}$ at each frame,  CTC implicitly models labels on sequence level. In the HMM framework, the label sequence is explicitly constrained and modeled by a n-gram language model.
  The sequence probabilities  in both frameworks are obtained by the forward-backward algorithm.
  \item {\em Confusion span}. $\tt{blank}$ is originally proposed in CTC to model the confusion between two label outputs discussed in Section~\ref{Sec:sgm-and-sdm}. Similar topologies are proposed  in Figure~\ref{fig:hmm-topo}(c-e). Beside topology differences, the $\tt blank$ in HMM framework is dependent on the sub-word unit, while  the CTC model shares a common $\tt blank$.
  \item {\em Competing hypothesis}. In word level CTC, the non-keyword elements are explicitly modeled by units, $\tt filler$ and $\tt wb$. In HMM, the competing hypotheses are implicitly modeled by sub-word units, and a sequence level prior probability, namely the sub-word level language model, is integrated in training stage to constrain the competing hypotheses. 
\end{itemize}
%


\section{Experiment}
\label{Sec:exp}

In the section, experiments are conducted on the proposed sequence discriminative training methods in both HMM framework and CTC framework. 
Both unrestricted KWS (spoken document retrieval task), and fixed vocabulary KWS (wakeup-word recognition task) are examined in the experiments. 
All experiments employed acoustic KWS based methods. The LVCSR based method is not included as previously discussed in Section~\ref{Sec:intro}. Moreover, a typical LVCSR based KWS method has a much larger acoustic  and language models, e.g.~\cite{meng2016non}. Thus the runtime overhead is much higher compared to the proposed system. 

\subsection{Spoken Document Retrieval in English}
\label{Sec:exp-sp-docu-detri}

\subsubsection{Experimental Setup}
\label{Sec:exp-sp-docu-detri-setup}

A speaker-independent 5k vocabulary dataset of the Wall Street Journal (WSJ0) corpus \cite{garofalo1993continous} was used to evaluate the proposed CTC lattice based KWS. Words or phrases which appear at least 5 times and whose length is between 3 and 12 phones were randomly selected as the keywords. In total, 50 keywords were used.

Because the size of WSJ0 is limited and the number of occurrences for each keyword is small,
both the HMM based systems and the CTC based systems are sub-word level, and the word level acoustic model will be discussed in Section~\ref{Sec:exp-wakeup-word-rec}.
The output labels are phones from the CMU pronunciation dictionary.
24-dimensional log filter-bank coefficients with their first and second derivatives at 10ms fixed frame rate were used as the input feature for all acoustic models. 
The setup of HMM based systems is similar to \cite{povey2016purely} but with less parameters as shown in Table~\ref{tab:model-discri} and \ref{tab:perf-all}~\footnote{As in \cite{povey2016purely}, the output frame rate is reduced by 3 folds.}. The phone level tri-gram language model with 36K n-grams is   estimated from the training set transcription. 
We use $\alpha=2.5$ and $\beta=2.5$ for NU-LF-bMMI.
The setup of the CTC model is the same to \cite{7736093}. Unidirectional LSTM is used in CTC due to online and low-latency requirements~\footnote{We have noticed some recent advances in latency-controlled BLSTM~\cite{ali-icassp17}, which might be the future work.}. 
LSTM is with 2 layers  each with 384 nodes. The projection layer has 128 nodes.
All the acoustic models are trained with Kaldi~\cite{povey2011kaldi}.

{\em Equal error rate} (EER) was taken as the sentence level
metric in the task, which reflects
the average error rate of the false alarm and false rejection. 
Lower EER is preferred. We also include  receiver operating characteristic (ROC) curves
in the summary of the experiment results.
In $\tt filler$ based decoding, denoted as {\em{kw-filler}}, the EER result was obtained by sweeping the transition probabilities between keyword and filler models. 
This transition probability can be regarded as the prior of keywords.
The thresholds of posterior smoothing, denoted as {\em{smooth}} and CTC hypothesis search, denoted as {\em{MED}}, are tuned based on threshold estimation algorithms in \cite{7736093}. Namely, 
\begin{equation}
\label{equ:eer-thres}
\begin{split}
\mathcal T_{EER}(\mathbf{k})=\mathcal T_0+\mathcal T(\mathbf{k})
\end{split}
\end{equation}
where $\mathcal T(\mathbf{k})$ is the estimated threshold of keyword sequence $\mathbf{k}$. $\mathcal T_0$ is shared for all keywords and  tuned  to  obtain the overall EER result.

{\em Real time factor} (RTF), the percentage of decoding time w.r.t.
the audio time, is taken to measure the overall efficiency of the model inference and the post-processing method. The lower RTF is the better.
In test stage, the machine setup is 
{\em{Intel(R) Xeon(R) CPU E5-2690 v2 @ 3.00 GHz}}. 

\subsubsection{HMM and CTC Model Training}
\label{Sec:exp-model-arch}
\begin{itemize}
\item {Generative sequence model.}
\end{itemize}

Empirical comparisons of the  acoustic model setups are conducted. 
Results on HMM based sequence discriminative training are shown in Table~\ref{tab:model-discri}.
All  models are trained by LF-MMI in Equation (\ref{equ:kws-mmi}), while using the kw-filler post-processing method. 
\begin{table}[thbp!]
  \caption{\label{tab:model-discri} {\it Model Architecture of  Sequence Discriminative Training for HMM}}
  \centerline{
    \begin{tabular}{c | c | c |c |c||c}
      \hline
      \multicolumn{1}{c|}{NN Model} &
      \multicolumn{1}{c|}{Context } &
      \multicolumn{1}{c|}{\# Param.} &
      \multicolumn{1}{c|}{CEW} &
      \multicolumn{1}{c||}{HMM} &
      \multicolumn{1}{c}{EER} \\
      \hline \hline
      \multirow{1}{0.1\columnwidth}{BLSTM}&\multirow{1}{0.05\columnwidth}{CD}& 0.60M & 0.1& \multirow{1}{0.05\columnwidth}{PB} & 3.3  
       \\
      \hline
      \hline
      \multirow{7}{0.1\columnwidth}{\textbf{TDNN}}&\multirow{1}{0.05\columnwidth}{CD} & 0.54M& 0.1&\multirow{1}{0.05\columnwidth}{PB}  & 3.3  
       \\
      \cline{2-6}
      &\multirow{6}{0.05\columnwidth}{\textbf{CI}}& \multirow{6}{0.1\columnwidth}{\textbf{0.51M}} & 0.1&\multirow{4}{0.05\columnwidth}{PB}  & 3.3  \\
      \cline{4-4}\cline{6-6}
      &&& 0.4&  & 3.2  \\
      \cline{4-4}\cline{6-6}
      &&&\textbf{0.7}& & 3.1  \\
      \cline{4-4}\cline{6-6}
      &&&1.0&& 3.2  \\
      \cline{4-5}\cline{6-6}
      &&&\multirow{2}{0.04\columnwidth}{0.7} &\multirow{1}{0.05\columnwidth}{\textbf{BP}}  & 3.0  \\
      \cline{5-5}\cline{6-6}
      &&&&\multirow{1}{0.05\columnwidth}{BPB}  & 3.0  \\
      \hline
    \end{tabular}
  }
\end{table}

The architecture of neural network is firstly examined. The first two rows compare bidirectional LSTM (BLSTM) and time delay neural network (TDNN) with similar number of  parameters~\footnote{Making the  number of parameters around 0.5M is suitable for general embedded applications. BLSTM is with 2 layers each with 80 nodes in both forward and backward layers. The projection layer is with 30 nodes. CD TDNN is with 7 layers each with 100 nodes and CI TDNN is with 7 layers each with 150 nodes.}. Result shows similar  EER performance. We believe it stems from: i) In KWS, the context needed for model inference is not very long, TDNN is enough for this application. ii) The number of model parameters is small in KWS, which also limits the performance of BLSTM. In the remaining experiment, only TDNN is taken due to its faster inference speed than BLSTM~\footnote{We have not tested LSTM in this framework yet. Recent progress has been made to combine the advantages of TDNN and LSTM~\cite{tdnnlstm}. This might be useful to improve KWS systems, which can be the future work.}. 
Secondly, the model unit is examined in the second and  the third row. The tri-phone state model, denoted as context dependent (CD), is compared with the mono-phone state model, denoted as context independent (CI).
The CD model
is based on three state left-to-right triphone models with 1536
tied states (senones). 
Its performance is similar to CI model~\footnote{\cite{povey2016purely} also shows similar trends. Besides, the performances of CD and CI model in BLSTMs are similar, i.e. EER=3.3\%. }. 
Thirdly, the cross-entropy regularization weight, denoted as CEW, is tuned in the third row  to the sixth row. The results show $0.7$ is the best, which is used for the remaining experiment. The reason is that the sub-word level language model is unavailable during test stage. Hence, the model should trade off between sequence discriminative and cross-entropy criteria. 
Finally, the topologies in Figure~\ref{fig:hmm-topo} are compared in the fifth, the seventh and the eighth rows. The proposed BP and BPB are both slightly better than PB proposed in \cite{povey2016purely}. 
However, statistical significance test on spotting results of 50 keywords is shown to be insufficient ($\alpha=0.05$, $p=0.18$).  
One explanation of the improvement can be from label delay~\cite{amodei2015deep} and  needs further research~\footnote{The experiment in HMM topologies is also conducted on larger dataset in LVCSR, and the current improvement is insignificant.}. 
Because the search space of BP is smaller than BPB while performances are similar, BP is preferred. 
The setup used in latter experiments is with bold-font in the table.

\begin{table}[thbp!]
  \caption{\label{tab:criteria-discri} {\it Criterion Comparison of HMM Based Discriminative Training }}
  \centerline{
    \begin{tabular}{c ||c}
      \hline
      \multicolumn{1}{c||}{Sequence Training Criterion } &
      \multicolumn{1}{c}{EER} \\
      \hline \hline
      LF-MMI &3.0 \\
      \textbf{LF-bMMI} &2.9 \\
      LF-sMBR &2.9 \\
      \hline\hline
      NU-LF-bMMI &2.7 \\
      \hline
    \end{tabular} 
  }
\end{table}
Different criteria proposed in Section~\ref{Sec:lfmmi-train} are compared in Table~\ref{tab:criteria-discri}. 
The post-processing method is kw-filler. Both LF-bMMI and LF-sMBR are slightly better than LF-MMI, which is similar to the conclusion in LVCSR~\cite{vesely2013sequence}. Since the training of LF-bMMI is faster than LF-sMBR and the speed is  comparable to LF-MMI, it is used in  later experiments~\footnote{The posterior calculation of bMMI criterion needs single pass of  forward-backward algorithm on each denominator lattice, while that of sMBR needs two passes as the implementation in~\cite{vesely2013sequence}.}. 
Besides, NU-LF-bMMI is also examined in this task, although it is originally proposed as a criterion for fixed vocabulary KWS in Section~\ref{Sec:exp-wakeup-word-rec-setup}. 
In this case, the 50 pre-defined keywords are used in the training stage to emphasize  gradients of keyword specific spans in the utterances.  Thus after NU-LF-bMMI training, the acoustic model is dependent on keywords.
It shows that NU-LF-bMMI brings about further improvement in the keyword-independent corpus, while the traditional method~\cite{chen2014small} proposed for fixed vocabulary KWS is not applicable for this corpus.   NU-LF-bMMI is not used for the remaining comparison, which is incomparable compared with other keyword independent systems. NU-LF-bMMI will be further examined in Section~\ref{Sec:exp-wakeup-word-rec}.


\begin{itemize}
\item {Discriminative sequence model.}
\end{itemize} 

Regarding to the sub-word level CTC, the introduction of the model unit $\tt wb$ is examined in Table~\ref{tab:wb}. Those keywords which contain less than 6 phones are considered as short keywords, otherwise the keywords are long.
Thus, the keyword set is divided into two parts to show the modeling effect clearly.
Both short keywords and long keywords yield performance improvements, namely 50\% relative EER reduction for short keywords and 42\%
EER reduction for long keywords. Besides, the performance
of long keywords consistently beats that of short keywords. This is because the phone sequences of short keywords are more likely
being substrings of other words or phrases, which will cause a
higher false alarm rate. By introduction of $\tt wb$, the problem is alleviated.

 \begin{table}[h]
 \caption{\label{tab:wb} {\it Performances of the Sub-word Level CTC with or without the $\tt wb$ Unit.}}
  \centerline
  {
\begin{tabular}{cc||c}
\hline 
Keyword Length & wb & EER \tabularnewline
\hline 
\hline 
\multirow{2}*{short} & $\times$ & 9.0 \tabularnewline
& $\bm{\surd}$ & 4.5 \tabularnewline
\hline\hline
\multirow{2}*{long} & $\times$ & 3.1 \tabularnewline
 & $\bm{\surd}$ & 1.8 \tabularnewline
\hline 
\end{tabular}
 }
\end{table}

\subsubsection{Post Processing and Speed Analysis}
\label{Sec:exp-post-process}

Different post-processing methods proposed for HMM framework in Section~\ref{Sec:post-process} and for  CTC framework in~\ref{Sec:post-process-ctc} are tested respectively. Note that the task in this section is an unrestricted KWS task, hence sub-word units are used to construct keywords. 

\begin{table}[thbp!]
  \caption{\label{tab:post-discri} {\it  Post Processing of Sequence Discriminative  Training Systems}}
  \centerline{
    \begin{tabular}{c |c||cc}
      \hline
      \multicolumn{1}{c|}{Model (Crit.) } &
      \multicolumn{1}{c||}{Post} &
      \multicolumn{1}{c}{EER} &
      \multicolumn{1}{c}{RTF}\\
      \hline \hline
      HMM&  smooth &9.8&0.008 \\
      (LF-bMMI)& \textbf{kw-filler} &2.9&0.028\\
      \hline\hline
       & smooth& 11.4  &0.026\\
     CTC & \textbf{kw-filler}&3.2&0.038\\
      & MED & 3.6 &0.031\\
      \hline
    \end{tabular}
  }
\end{table}
As table~\ref{tab:post-discri} shows, both in LF-bMMI and CTC, the posterior smoothing method obtains significantly worse performance than kw-filler. However, because its speed is much faster, the posterior smoothing method can be implemented as a pre-selection procedure of the speech corpus, and the kw-filler can be applied  as a verification procedure. Such strategies  improve the efficiency of spoken document retrieval systems, especially large amounts of data should always  be waded through by these systems. In CTC, the proposed MED method is also tested. It shows perceptibly worse performance than kw-filler, but better efficiency. The MED method will be further tested in the word level CTC in Section~\ref{Sec:exp-wakeup-word-rec}. 
The kw-filler system of CTC can be optimized by the phone synchronous decoding proposed in \cite{chen-is2016}, thus the search process in CTC is faster than that of LF-bMMI. The slower speed of the CTC system versus the LF-bMMI system stems from the model inference of LSTM for CTC is slower than that of TDNN for LF-bMMI.
%
%
%
%




\subsubsection{Performance Comparison}
\label{Sec:exp-perf-comp}

\begin{figure}
  \centering
    \includegraphics[width=\linewidth]{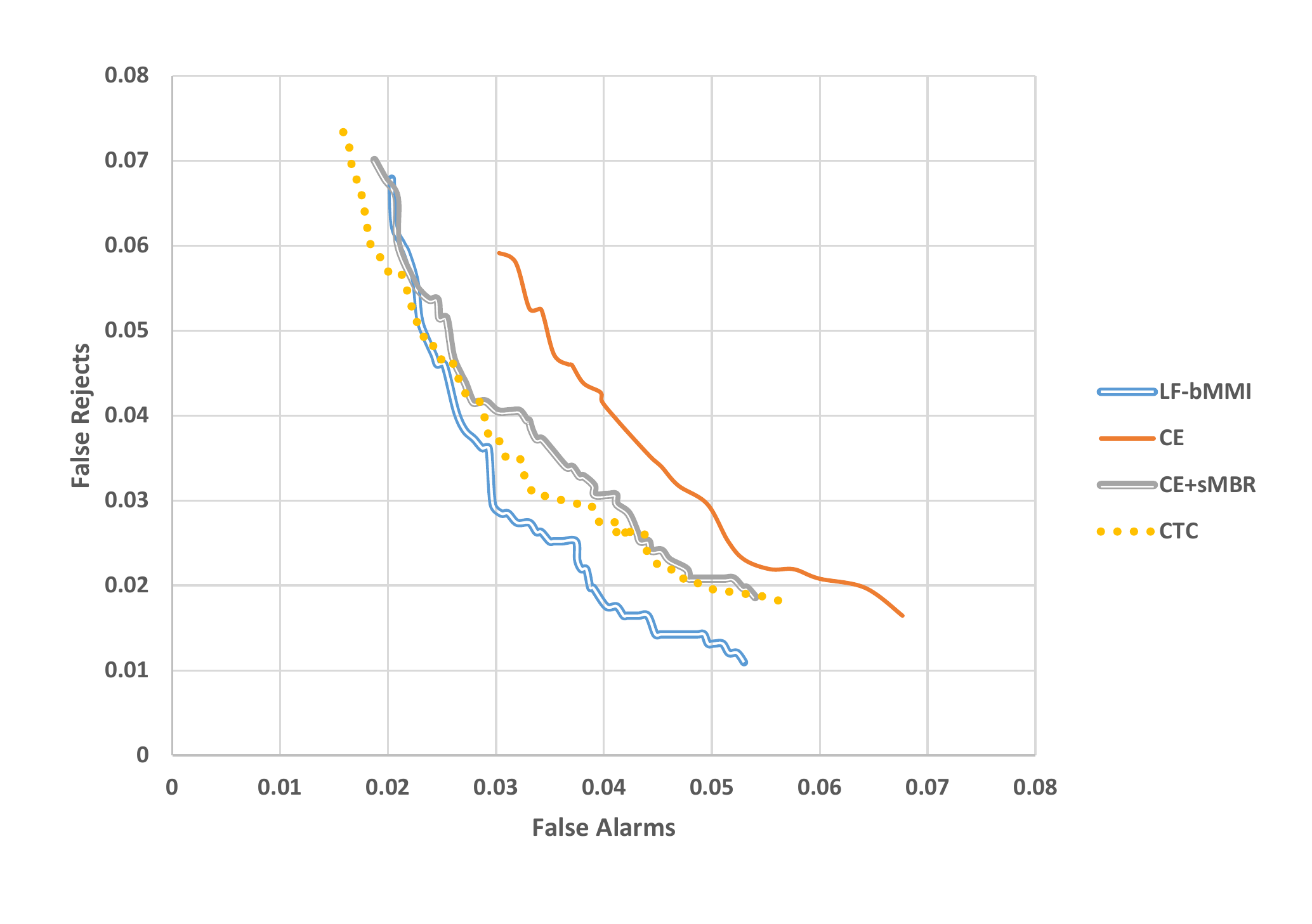}
    \vspace{-3em}
    \caption{\it ROC Curves Comparison in Unrestricted KWS. All systems used kw-filler as the post-processing approach.}
    \label{fig:roc}
\end{figure}

Finally, the performance and efficiency comparisons are summarized in 
Table~\ref{tab:perf-all}. Moreover, the ROC curves are shown in Figure~\ref{fig:roc}, where lower curves are better.
Note that all systems used the kw-filler post-processing approach.

\begin{table}[thbp!]
  \caption{\label{tab:perf-all} {\it  Performance and Efficiency Comparison in Unrestricted KWS. All systems used kw-filler as the post-processing approach.}}
  \centerline{
    \begin{tabular}{c | c | c |c ||cc}
      \hline
      \multicolumn{1}{c|}{Model } &
      \multicolumn{1}{c|}{Context } &
      \multicolumn{1}{c|}{\# Param. } &
      \multicolumn{1}{c||}{Criterion} &
      \multicolumn{1}{c}{EER} &
      \multicolumn{1}{c}{RTF} \\
      \hline \hline
     &\multirow{1}{0.05\columnwidth}{CD}&\multirow{1}{0.05\columnwidth}{0.6M}& CE &  4.0 & 0.051 \\
       {TDNN HMM }&\multirow{1}{0.05\columnwidth}{CD}&\multirow{1}{0.05\columnwidth}{0.6M}& CE+sMBR & 3.5 & 0.050
       \\
      &\multirow{1}{0.05\columnwidth}{CI}&\multirow{1}{0.05\columnwidth}{0.5M}& LF-bMMI &  \textbf{2.9} & \textbf{0.028}
       \\
       \hline\hline
       \multirow{1}{*}{LSTM CTC}&\multirow{1}{0.05\columnwidth}{CI}&\multirow{1}{0.05\columnwidth}{0.8M}& CTC & \textbf{3.2} & \textbf{0.038} \\
      \hline
    \end{tabular}
  }
\end{table}

The cross-entropy trained system is taken as the baseline in the first row.  It is a traditional NN-HMM based system with the clustered tri-phone state model unit. Namely the HMM topology is as depicted in Figure~\ref{fig:hmm-topo}(a).
Traditional discriminative training is then performed on the cross-entropy trained model. The lattice is generated by decoding the training dataset with a uni-gram language model estimated by the transcription as in~\cite{povey2007evaluation}.
The first two rows show that the traditional lattice based discriminative training with the original HMM topology relatively improves the performance by 12\%, which is similar to those for LVCSR~\cite{povey2005discriminative}. The slight improvement in RTF is because better acoustic model provides less confused hypotheses for the search process.

The third row versus the second row shows the  proposed best HMM based sequence discriminative training method compared with the traditional discriminative training method. 
There are 17\% relative improvement ($\alpha = 0.05$, $p = 0.06$, in statistical significance test) over  the traditional discriminative training  and 28\% relative improvement ($\alpha = 0.05$, $p = 10^{-5}$)  over the cross-entropy trained system. It is from better modeling effects in the proposed method~\footnote{We also conduct sMBR training over the model initialized by LF-bMMI and it shows very similar and inconsistent performance versus LF-bMMI, which is similar to~\cite{povey2016purely} in LVCSR.}:
\begin{itemize}
 \item As discussed in \cite{povey2016purely,pundak2016lower}, the modified HMM topology and lowered frame rate  both improve the performance. 
 \item The LVCSR lattice of the KWS model does not  contain good competing hypotheses for the traditional discriminative training. Specifically the lattice is recorded by 
decoding the KWS model with a word level language model. The decoding result of KWS model is always much worse than LVCSR models.
Thus worse competing hypothesis modeling deteriorates the performance.
 \item The word level language model used in the traditional discriminative training is unavailable in the test stage of KWS.  Nevertheless, LF-MMI uses a phone level language model, which is a good approximation for the lexicon used in the test stage.
\end{itemize}
Moreover, compared to the CE baseline, the LF-bMMI system almost doubles the speed. This stems from both the model inference and  search process. The former is mainly because of frame rate reduction. The latter is because the model unit of the proposed method is CI versus CD in the traditional method, which makes the search space much smaller. Notably, the {\tt filler} used in CD based systems is the clustered tri-phone states~\footnote{There are some efforts to improve efficiency of the {\tt filler} based KWS system by a further model unit selection~\cite{wang2010phonetic}, which haven't been included.}.

CTC system using LSTM is  provided in the last row, which also shows  better performance over the cross-entropy trained system ($\alpha = 0.05$, $p = 0.01$, in statistical significance test) and the traditional discriminative training system ($\alpha = 0.05$, $p = 0.11$). From Figure~\ref{fig:roc}, the CTC based method suffers more from false alarms, which is similar to the result in the next section.
Theoretically, the neural network takes the whole feature
sequence as input in CTC. Nevertheless, TDNN usually models the speech segments. Thus TDNN is not applied in CTC in this work. 
We also test BLSTM in CTC and the EER is $3.0$. However, the delay of BLSTM model is much longer than TDNN and LSTM. Thus the system is not listed in table.

Several traditional discriminative training methods for the  acoustic KWS based system discussed in Section~\ref{Sec:kws-and-disc} are not included in the comparison: i) They are not deep learning based system. ii) They are trained by  a keyword specific corpus. iii) Theoretical comparison is provided in Section~\ref{Sec:kws-and-disc} and the introduced  model units, e.g. {\tt filler}, in the proposed method is inspired by~\cite{sukkar1996utterance}. 
Besides, 
our preliminary trial on applying HMM based sequence discriminative training in CTC variant~\cite{sak2015fast,nict-icassp17} is not successful and shows no improvement: in word level CTC~\cite{li2018developing}, the LVCSR decoding lattice does not contain real competitors  (it only contains keywords and non-keyword units {\tt filler} inferred from the word level model)~\cite{sak2015fast}. In sub-word level model, as previously discussed, the LVCSR lattice of KWS model does not  contain good competing hypotheses for traditional discriminative training.

\subsection{Wakeup-word Recognition in Mandarin}
\label{Sec:exp-wakeup-word-rec}
\subsubsection{Experimental Setup}
\label{Sec:exp-wakeup-word-rec-setup}
In contrast to the previous section, a fixed vocabulary KWS task is used in this section. 
A Mandarin dataset similar to \cite{chen2014small} and \cite{cas-icassp17} is used in the training stage. It consists of two parts: general speech corpus and keyword specific corpus. 
The first part consists of 100 hours of spontaneous speech.
The second part consists of 30K positive examples of the keyword and 180K negative examples, namely phrases and utterances containing some substrings or similar pronunciations of the keyword. The positive examples were recorded by 50 males and 50 females with distance of 1, 3 and 5 meters in 5 different scenarios  respectively. In each case, the speaker needs to speak a keyword specific sentence three times with the speeds of fast, medium and slow. The real data were further augmented three folds to form the positive set. Operations include random speech concatenation and adding noise.

The test data  also consists of two parts: keyword specific set and environmental noise set. The first part aims to test the discrimination ability of the model to distinguish the keyword and the non-keyword speech as in \cite{chen2014small}.
It consists of 2K positive examples and 10K negative examples, representing 20\% of positive to negative ratio, to match expected application usage and recorded in varieties of scenarios as previously described. The previously proposed EER was taken as the metric.
The environmental noise set aims to  test the noise robustness of the model in rejecting keyword false alarm as in \cite{cas-icassp17}.
It consists of 300 hours of environmental noise recorded in varieties of scenarios. The number of false alarms per hour, denoted as {\em{false alarm frequency}} (FAF), was also taken as the metric. 
The hyper-parameters in post-processing are obtained from EER in the keyword specific set as Section~\ref{Sec:exp-sp-docu-detri-setup}. 
Each individual keyword holds its own specific threshold within $smooth$ method. The overall EER is the average  of the keyword EER.
Then all the hyper-parameters are fixed in the environmental noise set to obtain FAF.

In this task, 
both  word  and sub-word level systems are implemented.
HMM based systems and  CTC based systems are compared with the traditional cross-entropy trained system. 
The output label of the word level system is each word in the keyword sequence, and the output label of the sub-word level system is the no-tone syllable set in Mandarin.
The feature format and the setup of the acoustic model are the same as Section~\ref{Sec:exp-sp-docu-detri-setup}.
$\alpha=10.0$ and $\beta=10.0$ for NU-LF-bMMI.

\subsubsection{Results and Discussions}
\label{Sec:exp-wakeup-word-rec-result}


Table~\ref{tab:perf-mandarin} shows the result.
A cross-entropy trained TDNN model based on the word level model unit is taken as the baseline~\cite{chen2014small}. The post-processing is the traditional method, posterior smoothing.


\begin{table}[thbp!]
  \caption{\label{tab:perf-mandarin} {\it  Performance  and Efficiency Comparison in Fixed Vocabulary KWS}}
  \centerline{
    \begin{tabular}{c|m{4.5em}|c|c||ccc}
      \hline
      \multicolumn{1}{c|}{Model}
      &\multicolumn{1}{c|}{Unit} 
      &\multicolumn{1}{c|}{Criterion}&\multicolumn{1}{c||}{Post }&EER&FAF&RTF
      \\
      \hline\hline
      \cite{chen2014small}
      &\multirow{1}{*}{Word}
      &CE &smooth & 6.2 & 0.64 &0.014\\
     \hline\hline
      \multirow{3}{*}{HMM}&\multirow{3}{*}{Syllable}
      &CE & & 10.2 & 1.40 &0.041\\
      && LF-bMMI &kw-filler &8.3& 1.06 &0.033\\
      && NU-LF-bMMI& & \textbf{5.2} &\textbf{0.51} &\textbf{0.029}\\
      \hline\hline
      \multirow{2}{*}{CTC}
      &\multirow{1}{*}{Word}
      & \multirow{2}{*}{CTC} &smooth &9.1 & 1.13 & 0.024  \\
      &\ +{\tt filler}& &MED & \textbf{7.0}&\textbf{0.90} & \textbf{0.029}\\
      \hline
    \end{tabular}
  }
\end{table}

In the second row, the model is also CE-trained. The model unit is syllable and the post-processing is kw-filler.
The performance is deteriorated both in EER and FAF. The reason is that there are more model units in the sub-word level  system than those in the word level system. In the sub-word level system, all the model units in the syllable set are treated uniformly, while in the word level system, only the words as a substring of the keyword are included in the modeling.
Thus when obtaining EER in a keyword specific set, the false rejection rate of the former system is much worse, regardless of moderately better false alarm rate. Using the hyper-parameters obtained from EER evaluation, the system also performs worse in FAF.
The result is similar to \cite{chen2014small} and shows that  posterior smoothing based CE-trained system is a strong baseline.

In the third row, the proposed LF-bMMI system significantly improves the performance both in EER and FAF versus the cross-entropy trained system with the same model unit in the second row. However, it is still worse than the baseline in the first row. To overcome the false rejection problem arising from the uniform treatment of model units in the syllable set, NU-LF-bMMI is examined in the fourth row. To strengthen the discrimination ability of the acoustic model for specific
keywords, the gradients on them are weighted more significantly than those on non-keywords. Thus the performance regarding to the keyword specific set can be  improved.  Result shows that  NU-LF-bMMI significantly outperforms the word level cross-entropy trained system in the first row. We provide two reasons: Firstly, treating the keyword-related syllable units non-uniformly is theoretically similar to specifically model the  substrings of the keyword sequence as word level model units. Therefore, the modeling effects regarding to the keyword are similar. Secondly, the sub-word level system includes  the complete syllable set to model the non-keyword elements, which shows better performance in rejecting the false alarm caused  by both the non-keyword speech and the environmental noise. Regarding to the efficiency, although the model inference in NU-LF-bMMI is much faster than that in the cross-entropy trained system, NU-LF-bMMI additionally requires a  kw-filler search process, which results in nearly two folds of RTF versus the baseline with the posterior smoothing post-processing. Speeding up the search process in LF-MMI trained system is a future research topic.

Lastly, CTC based systems are compared. The system in the fifth row is similar to \cite{fernandez2007application}. Although achieving encouraging results in \cite{fernandez2007application}, the word level CTC based KWS system is worse than the baseline system. We believe the reason is that both the training and test dataset in our experiments are recorded in multiple conditions. It is more noisy and more difficult. Thus, the noise robustness of the CTC based system should be further improved to achieve better results. Specifically, the output distribution of CTC is fraughted with several false triggers, namely false alarms of keywords. 
The improvement of the  proposed method includes: Firstly, a specific unit $\tt filler$ to model the non-keyword speech is introduced, which improves both the keyword recognition ability and the context modeling between the keyword and the non-keyword speech. Secondly, the MED post-processing is proposed, which better utilizes the  sequence level competing hypothesis information by a two-stage filtering: lattice generation and hypothesis search in \cite{7736093}. The proposed method in the sixth row shows significantly better result compared with the traditional CTC-based system in the fifth row. However, the proposed CTC-based system, unlike in Section~\ref{Sec:exp-sp-docu-detri}, does not outperform the cross-entropy trained baseline. 
We suspect further analysis may focus on two directions: The noise robustness between different neural network architectures: TDNN, LSTM and BLSTM. Moreover, how to theoretically deal with the false triggers in CTC modeling.

\section{Conclusion}
\label{Sec:conclu}

The paper  improves deep learning based KWS by sequence discriminative training.
By introducing word-independent phone lattices or non-keyword blank symbols to construct competing
hypotheses, feasible and efficient sequence discriminative training
approaches are proposed for KWS.
Experiments are conducted on spoken document retrieval task and wakeup-word recognition tasks.
In the former, the vocabulary is always unrestricted, while in the latter, higher robustness is required (FAF in the experiment). Despite the distinct characteristics for each application,  
experiments showed  consistent and significant improvement,
compared to  previous frame-level deep learning approaches and separately optimized systems  in both fixed vocabulary and unrestricted KWS tasks.

Future works include three directions. i) Analyzing the noise robustness  between different sequence discriminative training frameworks~\cite{manohar2015semi,huang2013semi,chen2018progressive} and between different neural network architectures~\cite{qian2016neural,qian2016very,chen2018sequence}. ii)  Speeding up the search process in LF-MMI trained KWS systems~\cite{7736093,chen2018gpuwfst}. iii) Combining the recent advances in the confidence measure to improve the performance of the proposed KWS system~\cite{zhc00-chen-icassp17}.


%

%
%
%
%
\section*{Acknowledgment}

We thank the speech technology group of AISpeech Ltd. for infrastructure support and valuable technical discussions, Heinrich Dinkel for proofreading the manuscript.
\ifCLASSOPTIONcaptionsoff
  \newpage
\fi



%
%
%

\bibliographystyle{IEEEtran}

  \bibliography{mybib}
\end{document}